\documentclass[sigconf]{acmart}

\copyrightyear{2025}
\acmYear{2025}
\setcopyright{acmlicensed}
\acmConference[MM '25] {Proceedings of the 33rd ACM International Conference on Multimedia}{October 27--31, 2025}{Dublin, Ireland.}
\acmBooktitle{Proceedings of the 33rd ACM International Conference on Multimedia (MM '25), October 27--31, 2025, Dublin, Ireland}
\acmISBN{979-8-4007-2035-2/2025/10}
\acmDOI{10.1145/3746027.3755237}

\usepackage{multirow}
\usepackage{bm}
\usepackage{bbding}
\usepackage{tabularx}
\usepackage{enumitem}
\usepackage{comment}

\begin{document}
\def\mA{\mathcal{A}}
\def\mB{\mathcal{B}}
\def\mC{\mathcal{C}}
\def\mD{\mathcal{D}}
\def\mE{\mathcal{E}}
\def\mF{\mathcal{F}}
\def\mG{\mathcal{G}}
\def\mH{\mathcal{H}}
\def\mI{\mathcal{I}}
\def\mJ{\mathcal{J}}
\def\mK{\mathcal{K}}
\def\mL{\mathcal{L}}
\def\mM{\mathcal{M}}
\def\mN{\mathcal{N}}
\def\mO{\mathcal{O}}
\def\mP{\mathcal{P}}
\def\mQ{\mathcal{Q}}
\def\mR{\mathcal{R}}
\def\mS{\mathcal{S}}
\def\mT{\mathcal{T}}
\def\mU{\mathcal{U}}
\def\mV{\mathcal{V}}
\def\mW{\mathcal{W}}
\def\mX{\mathcal{X}}
\def\mY{\mathcal{Y}}
\def\mZ{\mathcal{Z}} 

\def\bbN{\mathbb{N}} 
\def\bbR{\mathbb{R}} 
\def\bbP{\mathbb{P}} 
\def\bbQ{\mathbb{Q}} 
\def\bbE{\mathbb{E}}

\def\1n{\mathbf{1}_n}
\def\0{\mathbf{0}}
\def\1{\mathbf{1}}

\def\A{{\bf A}}
\def\B{{\bf B}}
\def\C{{\bf C}}
\def\D{{\bf D}}
\def\E{{\bf E}}
\def\F{{\bf F}}
\def\G{{\bf G}}
\def\H{{\bf H}}
\def\I{{\bf I}}
\def\J{{\bf J}}
\def\K{{\bf K}}
\def\L{{\bf L}}
\def\M{{\bf M}}
\def\N{{\bf N}}
\def\O{{\bf O}}
\def\P{{\bf P}}
\def\Q{{\bf Q}}
\def\R{{\bf R}}
\def\S{{\bf S}}
\def\T{{\bf T}}
\def\U{{\bf U}}
\def\V{{\bf V}}
\def\W{{\bf W}}
\def\X{{\bf X}}
\def\Y{{\bf Y}}
\def\Z{{\bf Z}}

\def\a{{\bf a}}
\def\b{{\bf b}}
\def\c{{\bf c}}
\def\d{{\bf d}}
\def\e{{\bf e}}
\def\f{{\bf f}}
\def\g{{\bf g}}
\def\h{{\bf h}}
\def\i{{\bf i}}
\def\j{{\bf j}}
\def\k{{\bf k}}
\def\l{{\bf l}}
\def\m{{\bf m}}
\def\n{{\bf n}}
\def\o{{\bf o}}
\def\p{{\bf p}}
\def\q{{\bf q}}
\def\r{{\bf r}}
\def\s{{\bf s}}
\def\t{{\bf t}}
\def\u{{\bf u}}
\def\v{{\bf v}}
\def\w{{\bf w}}
\def\x{{\bf x}}
\def\y{{\bf y}}
\def\z{{\bf z}}

\def\balpha{\mbox{\boldmath{$\alpha$}}}
\def\bbeta{\mbox{\boldmath{$\beta$}}}
\def\bdelta{\mbox{\boldmath{$\delta$}}}
\def\bgamma{\mbox{\boldmath{$\gamma$}}}
\def\blambda{\mbox{\boldmath{$\lambda$}}}
\def\bsigma{\mbox{\boldmath{$\sigma$}}}
\def\btheta{\mbox{\boldmath{$\theta$}}}
\def\bomega{\mbox{\boldmath{$\omega$}}}
\def\bxi{\mbox{\boldmath{$\xi$}}}
\def\bnu{\mbox{\boldmath{$\nu$}}}                                  
\def\bphi{\mbox{\boldmath{$\phi$}}}
\def\bmu{\mbox{\boldmath{$\mu$}}}

\def\bDelta{\mbox{\boldmath{$\Delta$}}}
\def\bOmega{\mbox{\boldmath{$\Omega$}}}
\def\bPhi{\mbox{\boldmath{$\Phi$}}}
\def\bLambda{\mbox{\boldmath{$\Lambda$}}}
\def\bSigma{\mbox{\boldmath{$\Sigma$}}}
\def\bGamma{\mbox{\boldmath{$\Gamma$}}}
                                  
\newcommand{\myprob}[1]{\mathop{\mathbb{P}}_{#1}}

\newcommand{\myexp}[1]{\mathop{\mathbb{E}}_{#1}}

\newcommand{\mydelta}[1]{1_{#1}}

\newcommand{\myminimum}[1]{\mathop{\textrm{minimum}}_{#1}}
\newcommand{\mymaximum}[1]{\mathop{\textrm{maximum}}_{#1}}    
\newcommand{\mymin}[1]{\mathop{\textrm{minimize}}_{#1}}
\newcommand{\mymax}[1]{\mathop{\textrm{maximize}}_{#1}}
\newcommand{\mymins}[1]{\mathop{\textrm{min.}}_{#1}}
\newcommand{\mymaxs}[1]{\mathop{\textrm{max.}}_{#1}}  
\newcommand{\myargmin}[1]{\mathop{\textrm{argmin}}_{#1}} 
\newcommand{\myargmax}[1]{\mathop{\textrm{argmax}}_{#1}} 
\newcommand{\myst}{\textrm{s.t. }}

\newcommand{\denselist}{\itemsep -1pt}
\newcommand{\sparselist}{\itemsep 1pt}

\definecolor{pink}{rgb}{0.9,0.5,0.5}
\definecolor{purple}{rgb}{0.5, 0.4, 0.8}   
\definecolor{gray}{rgb}{0.3, 0.3, 0.3}
\definecolor{mygreen}{rgb}{0.2, 0.6, 0.2}

\newcommand{\cyan}[1]{\textcolor{cyan}{#1}}
\newcommand{\blue}[1]{\textcolor{blue}{#1}}
\newcommand{\magenta}[1]{\textcolor{magenta}{#1}}
\newcommand{\pink}[1]{\textcolor{pink}{#1}}
\newcommand{\green}[1]{\textcolor{green}{#1}} 
\newcommand{\gray}[1]{\textcolor{gray}{#1}}    
\newcommand{\mygreen}[1]{\textcolor{mygreen}{#1}}    
\newcommand{\purple}[1]{\textcolor{purple}{#1}}       

\definecolor{greena}{rgb}{0.4, 0.5, 0.1}
\newcommand{\greena}[1]{\textcolor{greena}{#1}}

\definecolor{bluea}{rgb}{0, 0.4, 0.6}
\newcommand{\bluea}[1]{\textcolor{bluea}{#1}}
\definecolor{reda}{rgb}{0.6, 0.2, 0.1}
\newcommand{\reda}[1]{\textcolor{reda}{#1}}

\def\changemargin#1#2{\list{}{\rightmargin#2\leftmargin#1}\item[]}
\let\endchangemargin=\endlist
                                               
\newcommand{\cm}[1]{}

\newcommand{\mhoai}[1]{{\color{magenta}\textbf{[MH: #1]}}}
\newcommand{\yifeng}[1]{{\color{red}\textbf{[yifeng: #1]}}}
\newcommand{\yftodo}[1]{{\color{blue}$\blacksquare$\textbf{[TODO: #1]}}}

\newcommand{\mtodo}[1]{{\color{red}$\blacksquare$\textbf{[TODO: #1]}}}
\newcommand{\myheading}[1]{\vspace{1ex}\noindent \textbf{#1}}
\newcommand{\htimesw}[2]{\mbox{$#1$$\times$$#2$}}


\newif\ifshowsolution
\showsolutiontrue

\ifshowsolution  
\newcommand{\Solution}[2]{\paragraph{\bf $\bigstar $ SOLUTION:} {\sf #2} }
\newcommand{\Mistake}[2]{\paragraph{\bf $\blacksquare$ COMMON MISTAKE #1:} {\sf #2} \bigskip}
\else
\newcommand{\Solution}[2]{\vspace{#1}}
\fi

\newcommand{\truefalse}{
\begin{enumerate}
	\item True
	\item False
\end{enumerate}
}

\newcommand{\yesno}{
\begin{enumerate}
	\item Yes
	\item No
\end{enumerate}
}

\newcommand{\Sref}[1]{Sec.~\ref{#1}}
\newcommand{\Eref}[1]{Eq.~(\ref{#1})}
\newcommand{\Fref}[1]{Fig.~\ref{#1}}
\newcommand{\Tref}[1]{Table~\ref{#1}}

\title{DualMat: PBR Material Estimation via\\ Coherent Dual-Path Diffusion}

\author{Yifeng Huang}
\orcid{0009-0003-9027-5963}
\authornote{Work initiated and partially completed during his internship at OPPO US Research Center. Email:\texttt{ yifehuang@cs.stonybrook.edu}}
\affiliation{%
  \institution{Stony Brook University}
  \city{Stony Brook}
  \state{NY}
  \country{USA}
}

\author{Zhang Chen}
\affiliation{%
  \institution{Meta}
  \city{Pittsburgh}
  \state{PA}
  \country{USA}
}

\author{Yi Xu}
\affiliation{%
  \institution{Goertek Alpha Labs}
  \city{Santa Clara}
  \state{CA}
 \country{USA}
}

\author{Minh Hoai}
\affiliation{%
 \institution{The University of Adelaide}
 \city{Adelaide}
 \state{SA}
 \country{Australia}
}
\author{Zhong Li$^\dagger$}
\affiliation{%
  \institution{Apple Inc.}
  \authornote{Corresponding author. Email: \texttt{zhonglee323@gmail.com}. The work was conducted when Zhong Li, Zhang Chen, Yi Xu was with OPPO.}
  \city{Cupertino}
  \state{CA}
  \country{USA}
}


\begin{abstract}
We present DualMat, a novel dual-path diffusion framework for estimating Physically Based Rendering (PBR) materials from single images under complex lighting conditions. Our approach operates in two distinct latent spaces: an albedo-optimized path leveraging pretrained visual knowledge through RGB latent space, and a material-specialized path operating in a compact latent space designed for precise metallic and roughness estimation. To ensure coherent predictions between the albedo-optimized and material-specialized paths, we introduce feature distillation during training. We employ rectified flow to enhance efficiency by reducing inference steps while maintaining quality. Our framework extends to high-resolution and multi-view inputs through patch-based estimation and cross-view attention, enabling seamless integration into image-to-3D pipelines. DualMat achieves state-of-the-art performance on both Objaverse and real-world data, significantly outperforming existing methods with up to 28\% improvement in albedo estimation and 39\% reduction in metallic-roughness prediction errors. Our project can be found at \href{https://yifehuang97.github.io/DualMatProjPage/}{yifehuang97.github.io/DualMatProjPage/}.
\end{abstract}



\maketitle

\section{Introduction}

Physically-Based Rendering (PBR) materials are fundamental to realistic computer graphics, defining how rendered surfaces should interact with light through properties such as diffuse reflection, metallic behavior, and surface roughness. These materials form the cornerstone of modern graphics applications, from video games and digital content creation to augmented/virtual reality (AR/VR). Accurate PBR material estimation is crucial for faithfully replicating real-world materials in digital environments, yet recovering these properties from a single image remains challenging due to the complex interplay of material attributes, lighting conditions, and underlying geometry. 

\begin{figure}[h]
\centering
\includegraphics[width=0.4\textwidth]{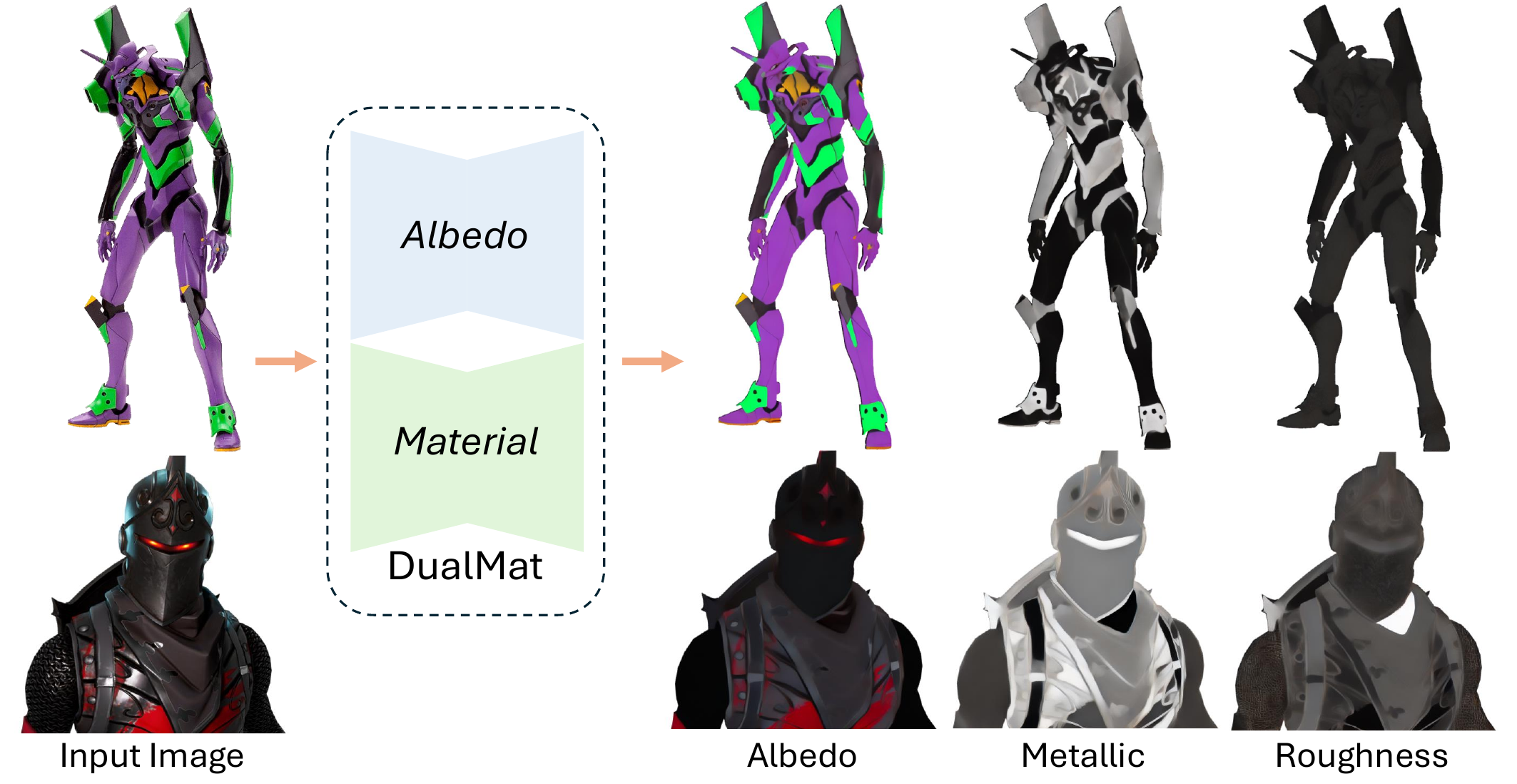}
\caption{Given a single input image (left), DualMat's dual-path design combines an albedo-optimized path and a material-specialized path to achieve accurate PBR material estimation.}
\label{fig:teaser}
\end{figure}

The importance of accurate PBR material estimation has been further amplified by recent breakthroughs in image-to-3D generation. Modern image-to-3D pipelines aim to reconstruct complete 3D assets from images. Early works like DreamFusion~\cite{poole2022dreamfusion} distill a 3D representation with NeRF~\cite{mildenhall2021nerf} from a 2D image diffusion model with Score Distillation Sampling loss. Recent works either employ large transformer architectures~\cite{hong2023lrm, li2023instant3d, wang2023pf, xu2024instantmesh, xu2024grm, xu2023dmv3d} to train a triplane representation in a data-driven manner, or explicitly generate multiview images and then reconstruct 3D models from the generated images~\cite{shi2023mvdream, liu2023syncdreamer, long2024wonder3d, shi2023zero123++, liu2023zero, li2024era3d,wang2023imagedream, tang2023make, wang2024crm}. However, these methods primarily focus on geometry and view-dependent appearance, often producing simplified or view-dependent textures that lack explicit material properties. This limitation severely restricts their applications in scenarios requiring physically accurate rendering, such as relighting under different environment maps or realistic integration into virtual environments. A robust PBR material estimation method could bridge this gap, enabling these 3D generation pipelines to produce assets with physically accurate rendering capabilities.

To address these challenges, PBR material estimation is crucial and can be formulated as an image-to-2D map translation task. Traditionally, this task has been tackled with methods heavily reliant on deep convolutional networks~\cite{vecchio2021surfacenet, li2017modeling, ye2018single, li2018materials, guo2021highlight, sang2020single, zhou2021adversarial, deschaintre2018single, deschaintre2019flexible, ye2021deep, guo2020materialgan, wen2022svbrdf, henzler2021generative, zhou2022tilegen, hu2022controlling, guo2020bayesian, liu2023openillumination, li2023relit}. While methods such as SurfaceNet have demonstrated the ability to estimate spatially-varying bidirectional reflectance distribution functions (SVBRDF) from a single image, these convolution-based architectures face significant limitations: they struggle to capture high-frequency details, resulting in blurred or oversimplified material maps; their performance degrades in multi-view scenarios where consistency across viewpoints is critical; and they lack the expressive power needed to generate coherent outputs under varying lighting conditions. The limitations of these conventional approaches suggest the need for a more powerful generative framework.

Recent advances in generative models, particularly diffusion models~\cite{ho2020denoising, ho2022cascaded, karras2022elucidating, rombach2022high, song2020score,liu2024panofree,xiong2025panodreamer}, offer promising solutions to these challenges. These models have demonstrated remarkable success in high-quality image synthesis~\cite{esser2024scaling, podell2023sdxl}, material generation~\cite{vainer2024collaborative, vecchio2024matfuse, vecchio2024controlmat, vecchio2024stablematerials, sartor2023matfusion}, super-resolution~\cite{saharia2022image, gao2023implicit, yue2024resshift}, and image translation~\cite{su2022dual, zhao2022egsde, sasaki2021unit}. Motivated by their ability to capture complex data distributions and generate high-fidelity outputs, we formulate PBR material estimation using a diffusion-based approach.

Building on these insights, we present DualMat, a novel dual-path diffusion framework for estimating PBR materials from single images, as shown in~\Fref{fig:teaser}. Our approach operates in two distinct latent spaces to effectively capture material properties and ensure consistency with input images. The albedo-optimized path leverages pretrained visual knowledge through RGB latent space to generate high-quality albedo maps, while the material-specialized path operates in a compact latent space designed to precisely model metallic and roughness properties.

To ensure coherent predictions between these complementary paths, we introduce a feature distillation loss during training that aligns intermediate representations. We further enhance computational efficiency by incorporating rectified flow, significantly reducing the number of denoising steps needed in both training and inference stages while preserving high-quality results.

Our framework extends beyond single-image estimation through two crucial adaptations: a patch-based strategy for high-resolution inputs that preserves fine details while maintaining global consistency, and a cross-view attention mechanism for multi-view scenarios that ensures consistent material estimation across viewpoints. These capabilities position DualMat as a versatile PBR estimator that seamlessly integrates into existing image-to-3D generation pipelines, enabling applications from realistic relighting to textured mesh generation.

In short, the primary contributions of our work are: (1) we propose a novel dual-path diffusion framework for PBR material estimation that leverages complementary latent spaces, consisting of an albedo-optimized path utilizing pre-trained visual knowledge and a material-specialized path dedicated to precise property estimation, combined with a feature distillation training strategy to ensure coherent outputs; (2) we design an efficient inference framework incorporating rectified flow sampling, patch-based high-resolution processing, and cross-view attention mechanisms, enabling high-quality material generation within just 2–4 steps and supporting seamless integration into image-to-3D pipelines; and (3) we demonstrate state-of-the-art performance across all evaluation metrics, achieving up to 28\% improvement in albedo estimation and a 39\% reduction in metallic-roughness prediction errors.

\section{Related Works}
This section reviews deep learning-based methods for material estimation (PBR or SVBRDF) and the application of diffusion models in material generation or estimation.

\begin{figure*}[ht]
   \centering
   \includegraphics[width=0.8\textwidth]{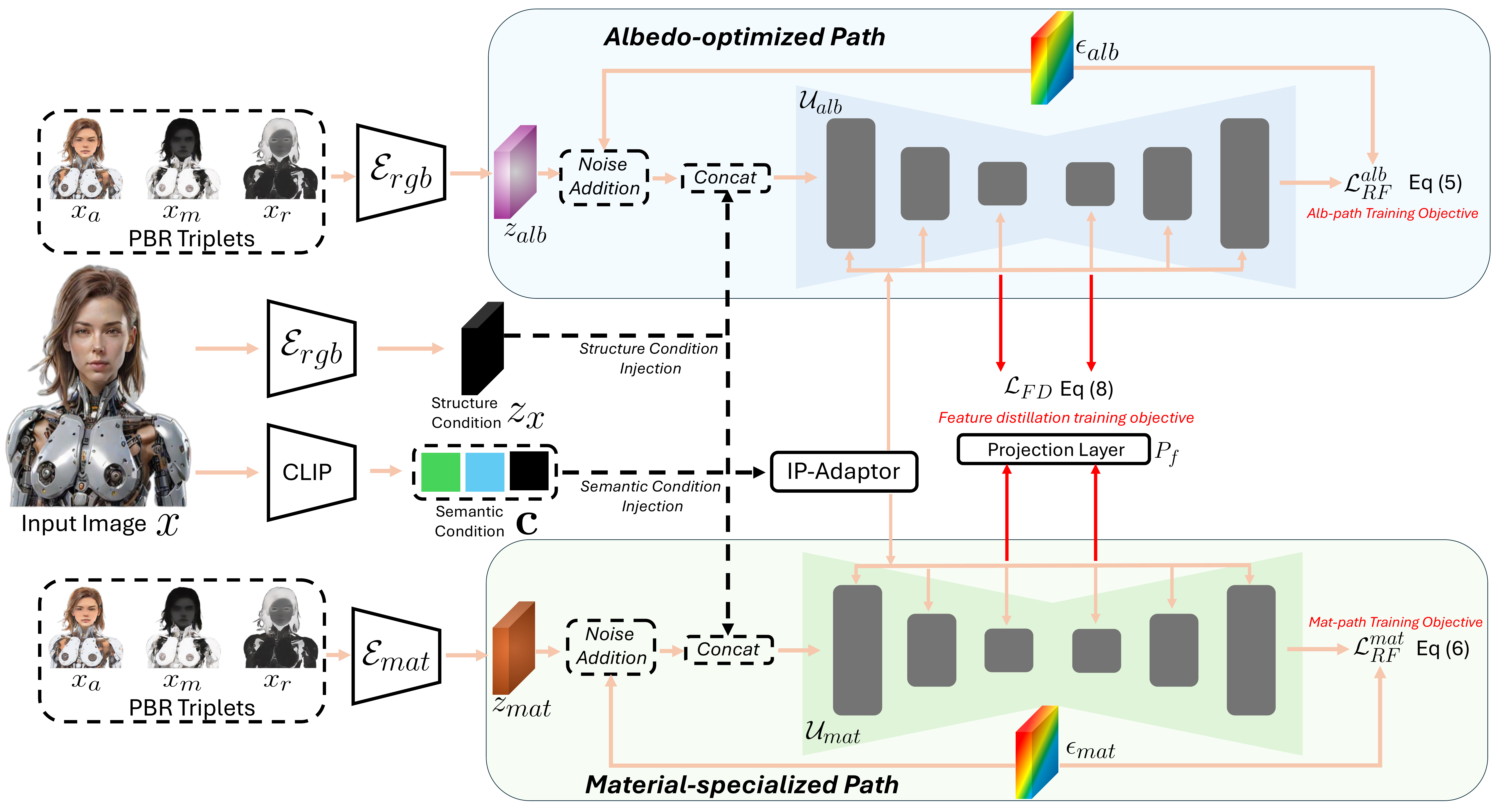}  
   \caption{DualMat's training architecture with dual paths and feature distillation. The albedo-optimized path (top) and material-specialized path (bottom) are trained with rectified flow objectives while maintaining consistency through feature distillation and dual conditioning mechanism.}
   \label{fig:framework}
\end{figure*}

\myheading{Material estimation.}  
Estimating spatially-varying material parameters, such as PBR and SVBRDF maps, from images is a well-established yet challenging task. Traditional methods leverage convolutional neural networks (CNNs) to capture material properties from single images under unknown natural lighting. Early CNN-based methods~\cite{li2017modeling, ye2018single} demonstrated plausible SVBRDF estimation for specific material classes, such as metals and plastics. Subsequent works \cite{li2018materials, guo2021highlight} introduced methods that recover SVBRDFs from flash-lit or casually captured images by incorporating physical priors. \citet{sang2020single} extended these approaches by jointly estimating material and shape with relighting capabilities. The emergence of GAN-based methods, such as SurfaceNet~\cite{vecchio2021surfacenet} and \citet{zhou2021adversarial}, enhanced detail recovery through adversarial training while bridging the synthetic-to-real gap using real image pairs. MaterialGAN~\cite{guo2020materialgan} and \citet{henzler2021generative} adapted GANs with learned latent spaces to serve as priors for inverse rendering and unsupervised SVBRDF reconstruction, respectively. \citet{wen2022svbrdf} improved material diversity by generating non-repeating BRDF fields from latent codes. TileGen~\cite{zhou2022tilegen} and \citet{hu2022controlling} further advanced GAN-based material modeling, generating tileable SVBRDFs with category-specific control and enabling texture transfer while preserving structural integrity. 

\myheading{Diffusion models in material estimation/generation.}  
The emergence of diffusion models has opened new possibilities for material estimation and generation. MatFusion~\cite{sartor2023matfusion} estimates SVBRDFs using diffusion, replacing residual convolution blocks with ConvNeXt~\cite{liu2022convnet} blocks and injecting image conditions by modifying the first convolution layer to accept additional input channels. \citet{vainer2024collaborative} model PBR material distributions with cross-network communication, linking a new PBR model to a frozen RGB model to tackle data scarcity and high-dimensional outputs. MatFuse~\cite{vecchio2024matfuse} offers a unified diffusion-based approach for 3D material creation and editing, using multiple conditioning sources and disentangled latent spaces for flexible map-level control. StableMaterials~\cite{vecchio2024stablematerials} applies semi-supervised latent diffusion, integrating adversarial training, a diffusion refiner, and a latent consistency model for fast, high-quality, and tileable material generation. Compared to prior CNN- and diffusion-based approaches, our dual-path diffusion framework captures high-quality material details while ensuring multi-view consistency through cross-view attention mechanisms, addressing challenges related to diverse lighting conditions and viewpoints.

\myheading{Image condition injection in diffusion.}  
A simple approach to inject image conditions is to concatenate them with the sampled noise~\cite{ke2024repurposing, saharia2022palette}. IP-Adapter~\cite{ye2023ip} refines this by extracting visual embeddings via a CLIP encoder, projecting them through a learned layer.

\section{Proposed Approach}

To introduce DualMat, we begin with preliminaries on diffusion models and rectified flow~\Sref{sec:prelim}, followed by PBR latent space design~\Sref{sec:pbr_latent}, dual-path diffusion model~\Sref{sec:dual_path_diffusion}, flow matching and feature transfer in training~\Sref{sec:flow_matching},  and extensions to high-resolution and multi-view scenarios~\Sref{sec:extensions}. Fig.~\ref{fig:framework}  illustrate our training pipelines.

\subsection{Preliminaries}
\label{sec:prelim}
\myheading{Diffusion models.} Diffusion models define a forward process that progressively corrupts data samples, transforming an initial data distribution $\mathbf{x}_0 \sim p_{\text{data}}(\mathbf{x})$ into a standard Gaussian distribution $\mathbf{x}_T \sim \mathcal{N}(\mathbf{0}, \mathbf{I})$ over $T$ time steps. The forward process follows a Markov chain, where at each step $t$, Gaussian noise is added according to: $\mathbf{x}_t \sim \mathcal{N}(\sqrt{\alpha_t}\mathbf{x}_{t-1}, (1-\alpha_t)\mathbf{I})$, with variance schedule $\alpha_t \in (0,1)$. To reverse this corruption, a neural network $f_\theta$ learns to predict the distribution $p_{\theta}(\mathbf{x}_{t-1} | \mathbf{x}_t)$ at each step. The training objective minimizes the prediction error of the added noise: $\mathbb{E}_{t, \mathbf{x}_0, \epsilon} \left[ \|\epsilon - \epsilon_\theta(\mathbf{x}_t, t)\|^2 \right]$, where $\epsilon$ represents the Gaussian noise from the forward process.

\myheading{Rectified flow.} Flow-based models offer an alternative approach by learning a continuous transformation between distributions through an ordinary differential equation. The Rectified Flow framework defines this transformation using $\frac{d\mathbf{x}_t}{dt} = v_\theta(\mathbf{x}_t, t)$, where $v_\theta$ represents a learned velocity field and $t \in [0,1]$. The model transforms data between the initial distribution $\mathbf{x}_0 \sim p_{\text{data}}(\mathbf{x})$ and a noise distribution $\mathbf{x}_1 \sim \pi_1(\mathbf{x})$ via linear interpolation: $\mathbf{x}_t = (1-t)\mathbf{x}_0 + t\mathbf{x}_1$. The training objective minimizes the flow matching loss:
\begin{equation}
    \min_\theta \mathbb{E}_{\mathbf{x}_1 \sim \pi_1, \mathbf{x}_0 \sim p_{\text{data}}} \left[ \int_0^1 \|(\mathbf{x}_1 - \mathbf{x}_0) - v_\theta(\mathbf{x}_t, t)\|^2 dt \right].  
\end{equation}
This formulation enables high-quality generation with significantly fewer steps than traditional diffusion models.

\subsection{PBR Latent Space Representation}
\label{sec:pbr_latent}
The foundation of our approach lies in establishing effective latent representations for PBR materials within the diffusion framework. We investigate two complementary encoding strategies that, while both capable of encoding complete PBR triplets, are designed to excel at different aspects of material estimation: (1) an albedo-optimized strategy leveraging pretrained image-based diffusion models, particularly Stable Diffusion, whose rich visual priors prove essential for high-quality albedo estimation despite a less compact representation; and (2) a material-specialized strategy developing a compact and efficient latent space, specifically optimized for precise metallic and roughness property estimation. This complementary design ultimately allows us to combine the strengths of both paths—superior albedo predictions from the albedo-optimized path and precise material properties from the material-specialized path.

\myheading{Albedo-optimized path via visual prior.} To leverage rich visual understanding from pretrained diffusion models for high-quality albedo estimation, we design our first path based on an established RGB encoder $\mathcal{E}_{rgb}$. This path processes PBR material components: albedo $(x_a)$, metallic $(x_m)$, and roughness $(x_r)$ through independent encoding. The albedo component, being inherently an RGB image, is processed directly through $\mathcal{E}_{rgb}$. For metallic and roughness components, which are single-channel maps, we employ channel repetition to satisfy the encoder's input specifications. The complete latent representation is formed by concatenating the individually encoded components:
\begin{equation}
z_{alb} = [\mathcal{E}_{rgb}(x_a), \mathcal{E}_{rgb}(R(x_m)), \mathcal{E}_{rgb}(R(x_r))],
\end{equation}
where $R(\cdot)$ represents our channel repetition operation that transforms single-channel inputs into three-channel representations compatible with $\mathcal{E}_{rgb}$. Specifically, utilizing the VAE architecture from Stable Diffusion v2~(SD2)~\cite{rombach2022high}, this encoding process yields a 12-channel latent vector, with four channels dedicated to each material component. While this approach effectively leverages pretrained visual understanding, particularly advantageous for high-quality albedo estimation, its independent processing of components may not fully capture the intricate interdependencies inherent in PBR materials.

\myheading{Material-specialized path with unified encoding.}
To achieve precise metallic-roughness estimation through joint material understanding, we design our second path using a specialized vector quantized architecture. Unlike the independent processing in the albedo-optimized path, this path employs a unified encoder $\mathcal{E}_{mat}$ that processes the complete PBR triplet $(x_a, x_m, x_r)$ simultaneously:
\begin{equation}
z_{mat} = \mathcal{E}_{mat}([x_a, x_m, x_r]),
\end{equation}
where $[x_a, x_m, x_r]$ represents the channel-wise concatenation of material components. Our architecture is specifically engineered to handle this 5-channel input (three for albedo, one each for metallic and roughness) and maps it into a compact 14-channel latent space. This unified encoding enables the capture of intricate relationships between material properties while maintaining a compact representation, crucial for precise metallic and roughness estimation. The encoder is trained with a comprehensive multi-objective loss:
\begin{equation}
\mathcal{L}_{mat} = \lambda_r\mathcal{L}_{rec} + \lambda_p\mathcal{L}_{perc} + \lambda_a\mathcal{L}_{adv} + \lambda_c\mathcal{L}_{code}.
\end{equation}
This loss consists of reconstruction $\mathcal{L}_{rec}$, perceptual $\mathcal{L}_{perc}$, adversarial $\mathcal{L}_{adv}$, and codebook commitment $\mathcal{L}_{code}$ terms to ensure high-quality material encoding.

\subsection{Dual-Path Diffusion for PBR Estimation}
\label{sec:dual_path_diffusion}
To synergistically leverage the rich visual understanding from pretrained diffusion models and specialized material modeling capabilities, we introduce DualMat, a novel dual-path diffusion framework illustrated in~\Fref{fig:framework}. While both paths are trained to predict complete PBR triplets $(x_a, x_m, x_r)$, each is architecturally optimized for different components: an albedo-optimized path operating in the pretrained visual latent space, and a material-specialized path working in our unified compact latent space.

Each path employs a dedicated denoising U-Net ($U_{alb}$ and $U_{mat}$) carefully tailored to its respective latent representation. This dual-path design enables specialization where it matters most---the albedo-optimized path excels at generating high-fidelity albedo predictions by leveraging pretrained visual knowledge, while the material-specialized path achieves precise metallic and roughness estimations through its unified material understanding. Although both paths produce complete PBR predictions, we strategically combine their strengths in the final output: taking the albedo component from $U_{alb}$ and the metallic-roughness components from $U_{mat}$, formally expressed as $(x_a^{alb}, x_m^{mat}, x_r^{mat})$. The consistency between these paths is ensured through feature distillation during training, as detailed in the following sections.

To ensure robust conditioning on the input image $x$, DualMat incorporates two complementary conditioning mechanisms that operate at different levels of abstraction:

1) \textbf{High-level semantic conditioning.} We leverage the rich semantic understanding of CLIP through IP-Adapter. Specifically, we extract deep visual features $\mathbf{c} = P_{ip}(\mathcal{E}_{clip}(x))$, where $\mathcal{E}_{clip}$ is the CLIP vision encoder and $P_{ip}$ represents the IP-Adapter projection. These semantic features are integrated via cross-attention layers throughout both U-Nets, enabling a global understanding of material properties for each path's specialized focus.

2) \textbf{Low-level structure conditioning.} We further strengthen the connection to input image details through direct latent conditioning $z_x = \mathcal{E}_{rgb}(x)$. By concatenating this latent representation with the noise input at each denoising step, we provide fine-grained structural guidance that preserves local material details and ensures spatial coherence across predictions.

\subsection{Flow Matching and Feature Transfer}
\label{sec:flow_matching}
To achieve both efficient training and consistent predictions across paths, we introduce two key mechanisms in our framework. First, we adopt rectified flow to enable rapid inference, modeling the continuous transformation between data and noise distributions with remarkably few sampling steps (2--4) compared to traditional diffusion approaches. Second, given that our final prediction combines albedo from the albedo-optimized path $(x_a^{alb})$ with material properties from the material-specialized path $(x_m^{mat}, x_r^{mat})$, we incorporate feature transfer strategies to ensure cross-path consistency. In our dual-path framework, each U-Net predicts a velocity field that guides the noise-to-data transformation:
\small
\begin{align}
\mathcal{L}_{RF}^{alb} &= \mathbb{E}_{t, x, \epsilon_{alb}} \left[\left\|(z_{alb} - \epsilon_{alb}) - v_{\theta_{alb}}(z_t^{alb}, t, x)\right\|^2\right],  \\
\mathcal{L}_{RF}^{mat} &= \mathbb{E}_{t, x, \epsilon_{mat}} \left[\left\|(z_{mat} - \epsilon_{mat}) - v_{\theta_{mat}}(z_t^{mat}, t, x)\right\|^2\right], 
\end{align}
\normalsize
where $z_{alb} = \mathcal{E}_{rgb}(x)$ and $z_{mat} = \mathcal{E}_{mat}(x)$ encode the PBR materials in their respective latent spaces, $v_{\theta}$ represents the predicted velocity field, $z_t$ denotes the noisy latent at timestep $t$, and $x$ is the input image. These velocity fields guide the progressive refinement from noise to structured material representations.

Our training pipeline consists of two strategically designed stages to ensure consistent predictions while maintaining path specialization:

1) \textbf{Albedo path training:} We initialize $U_{alb}$ from a pretrained Stable Diffusion v2 model and finetune it with $\mathcal{L}_{RF}^{alb}$. The model's first and last convolutional layers are carefully adapted for PBR material estimation while preserving the rich visual understanding encoded in the pretrained weights, essential for high-quality albedo prediction.

2) \textbf{Material path training:} With $U_{alb}$ frozen, we train $U_{mat}$ from scratch using both $\mathcal{L}_{RF}^{mat}$ and feature distillation. This distillation transfers crucial visual knowledge from the albedo-optimized path while ensuring consistency between material property predictions and albedo estimates.

For effective feature transfer, we align the intermediate representations across both paths through a learned projection $P_f$ as shown in \Fref{fig:framework}. At each network layer $l$, we minimize the $L_2$ distance between albedo path features and their projected material path counterparts:
\begin{equation}
\mathcal{L}_p^l = \left\| f_l^{alb} - P_f(f_l^{mat}) \right\|_2^2,
\end{equation}
where $f_l^{alb}$ and $f_l^{mat}$ denote feature maps from corresponding layers. The total feature distillation loss aggregates these alignments across the entire U-Net hierarchy:
\begin{equation}
   \mathcal{L}_{FD} = \sum_{l \in \mathcal{L}} \left\| f_l^{alb} - P_f(f_l^{mat}) \right\|_2^2,
\end{equation}
where $\mathcal{L}$ spans four down-sampling blocks, one middle block, and four up-sampling blocks. This comprehensive feature alignment ensures that while each path maintains its specialized capabilities, albedo prediction in $U_{alb}$ and material property estimation in $U_{mat}$, their predictions remain coherent when combined in the final output $(x_a^{alb}, x_m^{mat}, x_r^{mat})$.

\subsection{High-Resolution and Multi-View Support}
\label{sec:extensions}
\myheading{High-resolution processing.} While our model is trained at fixed resolutions (e.g., 256, 512), real-world applications often require processing higher-resolution images. Simply downsampling the input would result in loss of fine details, so we propose a coarse-to-fine strategy that maintains high-frequency details while ensuring global-local consistency.

First, we obtain a global image prediction $I^{g}$ by applying our model to a
down-sampled input. We then process the full-resolution image in overlapping
patches that match the training resolution. During patch-wise inference, we
blend global and local guidance: 
\small
\begin{equation}
\hat{z}_{t}=z_{t}
+\gamma\nabla_{z_{t}}\!\Bigl(
      \|D(z_{0})_{\text{blur}}-I^{g}_{p}\|_{2}^{2}
      +\!\!\!\sum_{n\in\{l,t\}}
        \|D(z_{0})_{p_{n}}-I^{n}_{p_{n}}\|_{2}^{2}
\Bigr),
\end{equation}
\normalsize
where $z_{0}$ is the predicted clean latent, $D(\cdot)$ decodes a latent to
RGB image space, $I^{g}_{p}$ is the aligned patch from the global prediction,
and $I^{n}_{p_{n}}$ are the overlapping regions from the previously generated
left $(n=l)$ and top $(n=t)$ patches. This image-space formulation preserves
fine details while ensuring seamless boundaries between windows.

\myheading{Multi-view consistency.} 
To ensure consistent PBR material estimation across multiple views, which is essential for effective integration into image-to-3D generation tasks, we enhance our single-view diffusion models by incorporating cross-view attention mechanisms. Specifically, we fine-tune both the albedo-optimized and material-specialized paths, initially trained for single-view predictions, using multi-view inputs (e.g., four-view images) during an additional fine-tuning stage.

During this process, each self-attention layer within our denoising U-Nets ($U_{alb}$ and $U_{mat}$) is modified to perform attention operations across tokens from all viewpoints simultaneously. Given multiple input views, we first independently encode each view into its latent representation \( z^{(v)} \). These latent tokens from all views are concatenated into a single sequence as:
$
z = [z^{(1)}, z^{(2)}, \dots, z^{(V)}],
$
where \( V \) denotes the total number of views. The concatenated tokens are then processed using a self-attention mechanism:
\[
\text{Attention}(z) = \text{softmax}\left(\frac{Q(z)K(z)^T}{\sqrt{d}}\right)V(z),
\]
where \( Q(\cdot) \), \( K(\cdot) \), and \( V(\cdot) \) represent query, key, and value projections, respectively, and \( d \) is the dimension of attention heads. After this cross-view attention operation, the tokens are separated back into individual view representations:
$
[z^{(1)'}, z^{(2)'}, \dots, z^{(V)'}] = \text{Split}(\text{Attention}(z)).
$

This cross-view attention enables each view's latent representation to interact and integrate relevant information from other viewpoints, thereby promoting coherent and aligned material properties across all views. Fine-tuning our diffusion models with this mechanism significantly enhances multi-view consistency.

\section{Experiments}
\label{sec:Exp}

\begin{figure*}[ht]
\centering
\includegraphics[width=0.9\textwidth]{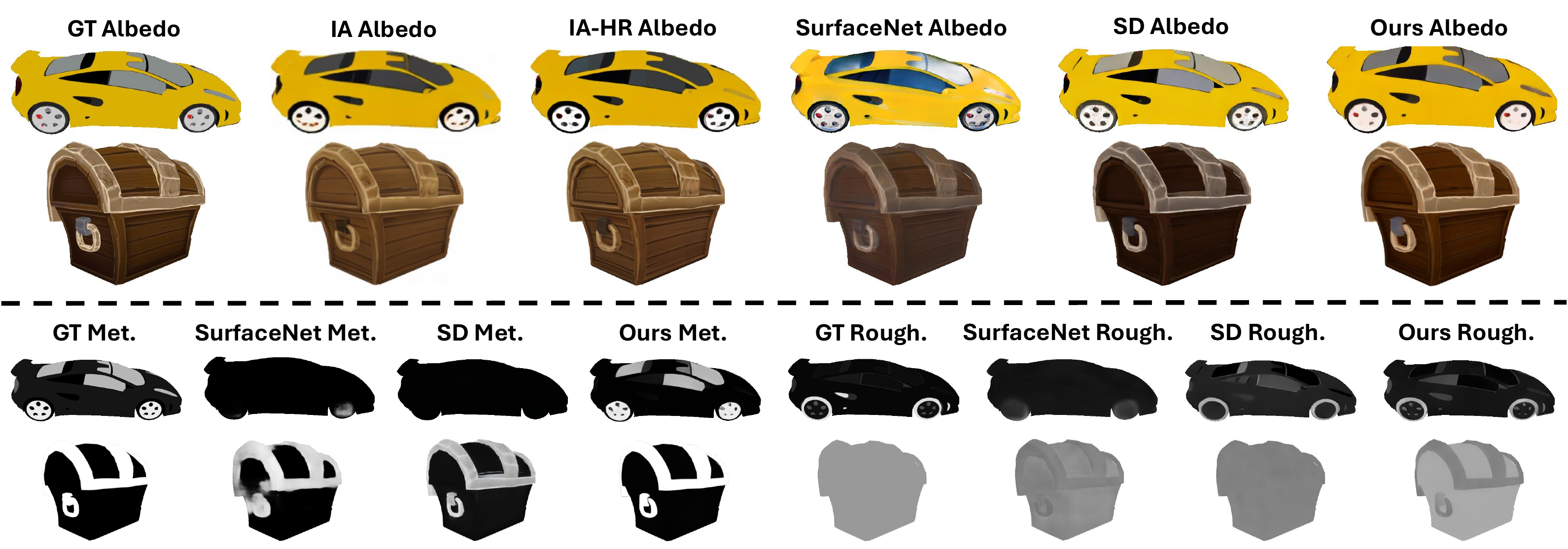}
\caption{Comparison on Objaverse test objects, illustrating material decomposition capabilities of our method versus baseline approaches.}
\label{fig:qual_obja}

\includegraphics[width=0.9\textwidth]{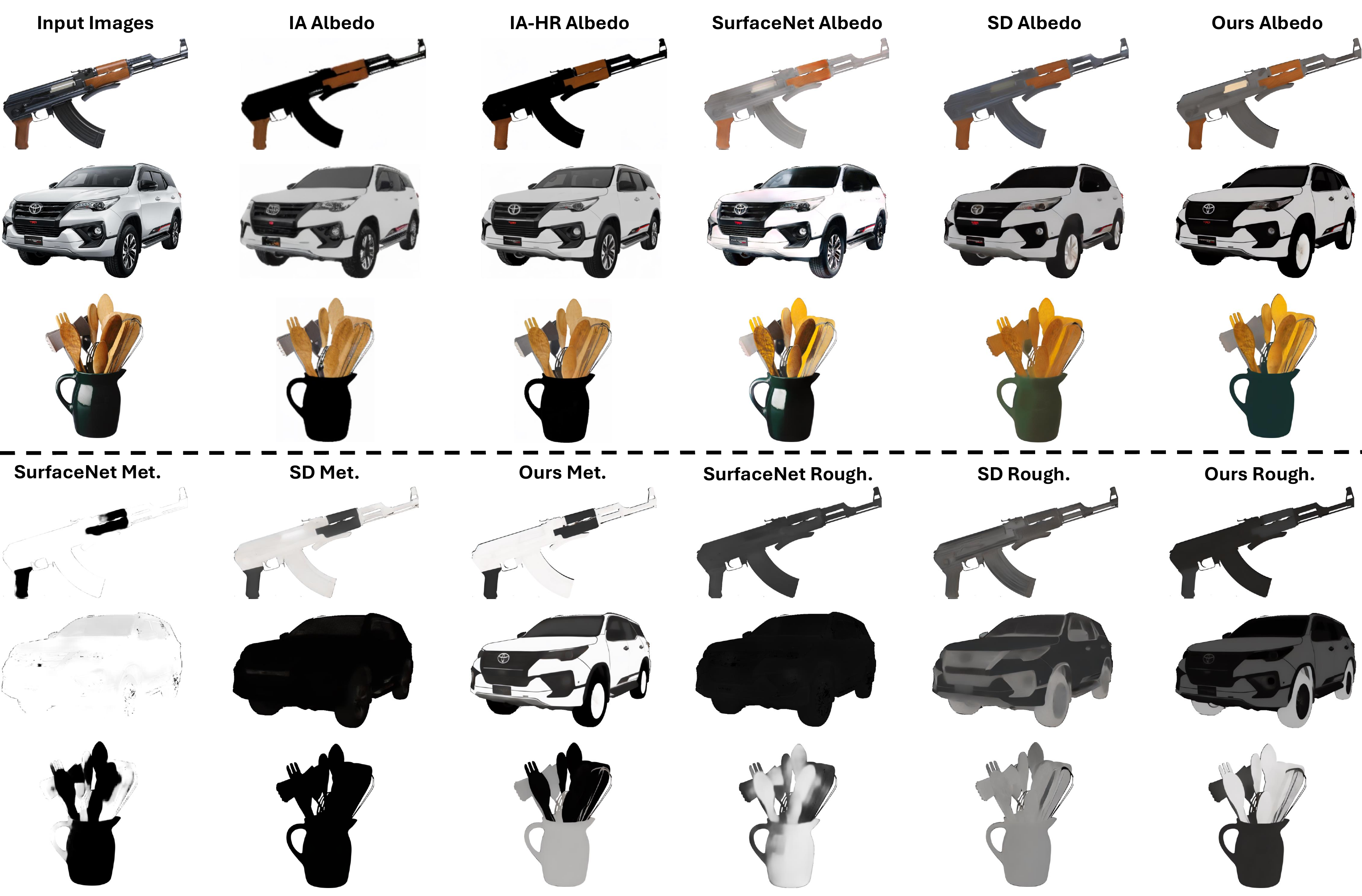}
\caption{Results on real-world images, demonstrating our method's superior generalization to diverse real-world materials with improved detail preservation and accurate material separation.}

\label{fig:qual_wild}
\end{figure*}

\subsection{PBR Estimation} 

\myheading{Dataset.} We construct our training dataset from Objaverse by selecting 65,000 objects with valid PBR materials. Each object is rendered from 16 different viewpoints under 100 distinct HDR environment maps, yielding over one million image-PBR pairs for training.

\myheading{Evaluation protocol.} We evaluate our method primarily on the fundamental task of single-view material estimation. Using a test set of 654 objects from Objaverse (rendered under novel HDR environments), we assess each view independently, even when multiple views of the same object are available. For each single-view estimation, we employ multiple quality metrics: PSNR, SSIM, and LPIPS for albedo evaluation, and RMSE for metallic and roughness components. This single-view evaluation protocol provides a rigorous assessment of our method's core material estimation capabilities, separate from its multi-view extensions.

\myheading{Baselines.} We compare DualMat against three representative methods: Intrinsic Anything~\cite{chen2024intrinsicanything} (trained on Objaverse, albedo estimation only), SurfaceNet~\cite{vecchio2021surfacenet} (fine-tuned with PBR prediction head), and Stable Diffusion (adapted with extended convolution layers). For a fair comparison, both SurfaceNet and Stable Diffusion are trained on our dataset with identical settings.

\myheading{Implementation details.}
Our material-specialized encoder $\mathcal{E}{mat}$ is trained for 1.5M iterations with a batch size of 32. The loss weights are set as follows: reconstruction (1.0), LPIPS (0.001), commitment (0.1), and adversarial (0.01). The albedo-optimized U-Net ($U{alb}$) is fine-tuned from Stable Diffusion v2 (SD2) for 500K iterations with a batch size of 64. Subsequently, we train the material-specialized U-Net ($U_{mat}$) from scratch for 500K iterations with a batch size of 48, employing both rectified flow and feature distillation objectives. Notably, the feature distillation loss is applied only during the initial 50K iterations. Both U-Nets share identical architectures except for their input and output layers. During inference, we achieve high-quality results using only three sampling steps. Additionally, we preserve the text-conditioning architecture with empty embeddings to facilitate potential future extensions. All experiments are conducted using four NVIDIA A100 GPUs.

\myheading{Results.} The quantitative evaluation results are summarized in~\Tref{tab:quan_result}. DualMat consistently outperforms existing methods across all metrics. For albedo estimation, our method achieves significant improvements in both pixel-wise accuracy (PSNR: 28.6 dB, SSIM: 0.932) and perceptual quality (LPIPS: 0.047), surpassing Intrinsic Anything even with its high-resolution guidance enhancement (PSNR: +4.4 dB, LPIPS: -0.026). The dual-path design proves particularly effective for metallic and roughness estimation, where we achieve RMSE values of 0.057 and 0.060, respectively, representing a 39\% reduction in estimation error compared to previous methods.

\begin{table}[h!]
\centering
\begin{tabular}{lcccccc}
\toprule
Method & \multicolumn{3}{c}{Albedo} & \multicolumn{1}{c}{\multirow{2}{*}{Met.}} & \multicolumn{1}{c}{\multirow{2}{*}{Rough.}} \\ 
\cmidrule(lr){2-4}
& PSNR & SSIM & LPIPS & & \\
\midrule
IA~\cite{chen2024intrinsicanything}    & 23.4 & 0.895 & 0.092 & - & - \\
IA-HR$^\dagger$~\cite{chen2024intrinsicanything}  & 24.2 & 0.912 & 0.073 & - & - \\
SurfaceNet~\cite{vecchio2021surfacenet} & 26.1 & 0.931 & 0.052 & 0.093 & 0.092 \\
SD & 26.4 & 0.918 &  0.053 & 0.089 & 0.089 \\
DualMat(Ours) & \textbf{28.6} & \textbf{0.932} & \textbf{0.047} & \textbf{0.057} & \textbf{0.060} \\
\bottomrule
\end{tabular}
\caption{Quantitative evaluation on single-view PBR material estimation. We report PSNR, SSIM, and LPIPS for albedo quality, and RMSE for metallic (Met.) and roughness (Rough.) components. Our method achieves superior performance across all metrics on the Objaverse test set. $^\dagger$IA-HR refers to Intrinsic Anything with high-resolution guidance enhancement.}
\label{tab:quan_result}
\vspace{-0.5cm}
\end{table}

\subsection{Qualitative Results}
\begin{figure}[H]
  \centering
  \includegraphics[width=\columnwidth]{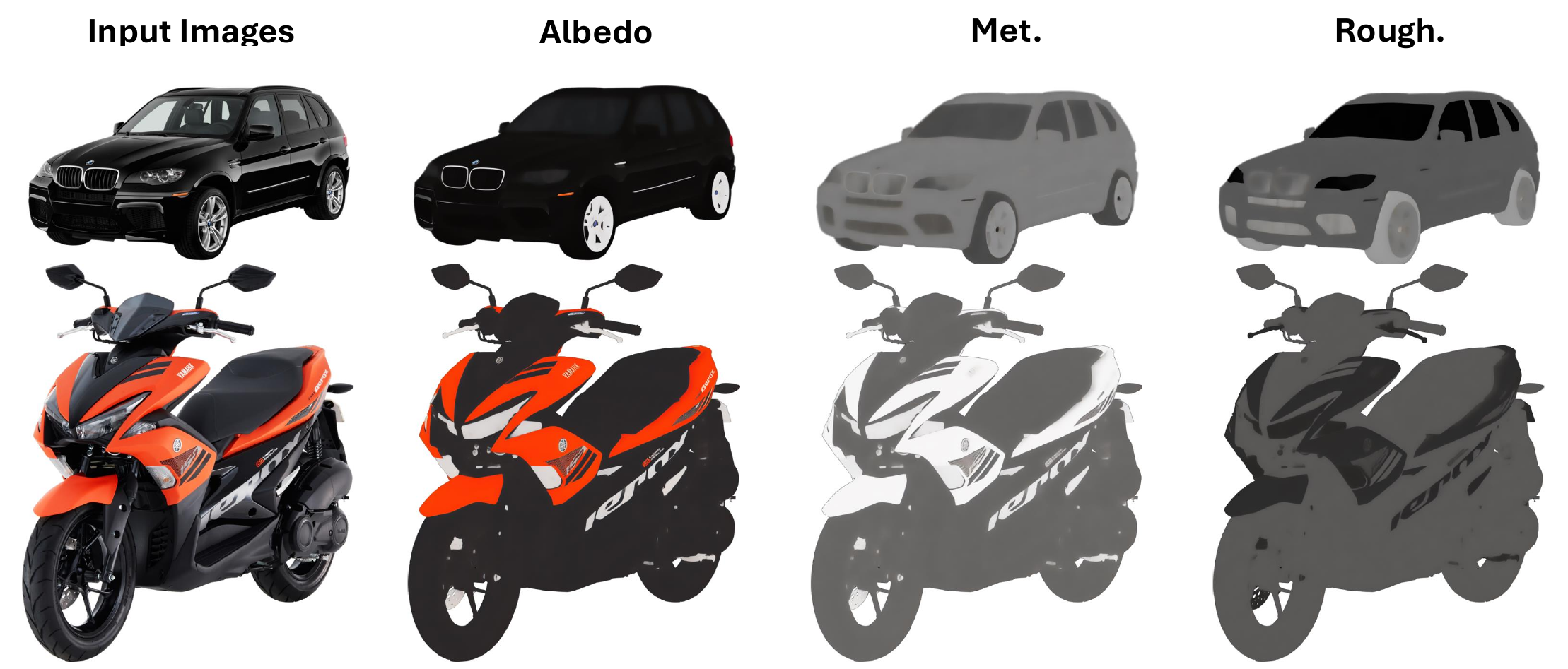}
  \caption{Qualitative results on highly specular objects.}
  \label{fig:qual_high_spec}
\end{figure}

\Fref{fig:qual_obja} demonstrates our method's qualitative performance on Objaverse test objects, where our method shows improvements in material decomposition over existing approaches. Most notably, in \Fref{fig:qual_wild} with real-world objects, our method demonstrates significantly better generalization capability, preserving high-frequency details such as car logos and achieving cleaner albedo-metallic separation in challenging materials like car windows. These results validate our method's effectiveness in handling complex real-world materials while maintaining intricate surface details. Additional qualitative results for highly specular objects are provided in~\Fref{fig:qual_high_spec}.

\subsection{Ablation Study} 
\subsubsection{Ablation on Dual-Path Architecture}
We conduct ablation studies to validate the necessity of our dual-path architecture, as shown in \Tref{tab:ABA_Conponent}, comparing three variants: using only the material-specialized path, only the albedo-optimized path, and our full dual-path model. The results demonstrate clear complementary benefits of both paths. The albedo-optimized path alone achieves strong albedo quality (matching our dual-path results in PSNR and LPIPS) but struggles with material property estimation. Conversely, the material-specialized path provides precise metallic-roughness estimates (matching our dual-path performance) but lower albedo quality. Our dual-path design successfully preserves the optimal components from each path, maintaining both high albedo quality and precise material property estimation.

\begin{table}[t]
\centering
\begin{tabular}{lccc}
\toprule 
$\mathcal{U}_{mat}$ & \CheckmarkBold& \XSolidBrush& \CheckmarkBold \\
$\mathcal{U}_{alb}$ & \XSolidBrush& \CheckmarkBold& \CheckmarkBold \\
\midrule 
Albedo PSNR& 27.8 & 28.6 & \textbf{28.6}\\
Albedo SSIM& 0.928 & 0.932 & \textbf{0.932}\\ 
Albedo LPIPS& 0.056 & 0.047 & \textbf{0.047}\\ 
Metallic RMSE& 0.057 & 0.065 & \textbf{0.057}\\ 
Roughness RMSE& 0.061 & 0.064 & \textbf{0.060}\\ 
\bottomrule 
\end{tabular}
\caption{Ablation study comparing single-path versus dual-path performance. Our dual-path design achieves optimal performance across all metrics.}
\label{tab:ABA_Conponent}
\end{table}

\subsubsection{Ablation on Multi-view Consistency}
\begin{figure}[H]
\centering
\includegraphics[width=\columnwidth]{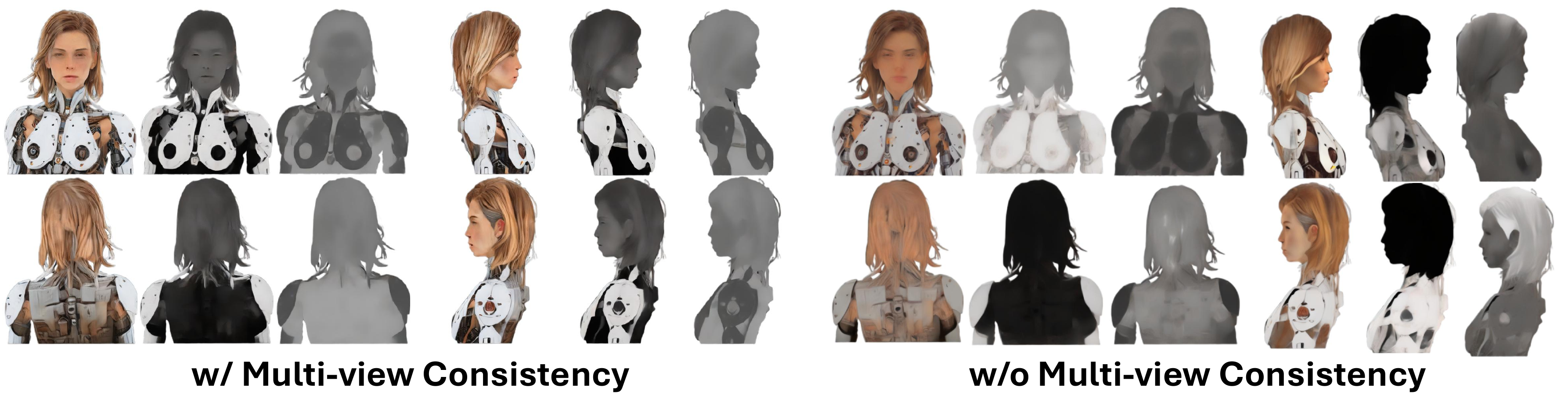}
\caption{Qualitative ablation study demonstrating the effectiveness of our proposed multi-view consistency mechanism. }
\label{fig:aba_multi_view}
\end{figure}

We present a qualitative ablation in \Fref{fig:aba_multi_view}, demonstrating the impact of our multi-view consistency mechanism. With multi-view consistency enabled, our model significantly enhances the coherence and alignment of predicted material maps across different viewpoints. As illustrated in the comparison, omitting multi-view consistency introduces noticeable artifacts and inconsistencies within the albedo, metallic, and roughness maps, particularly around intricate structures and boundaries. In contrast, incorporating our proposed cross-view attention mechanism ensures stable and coherent predictions, effectively preserving fine surface details and maintaining visual consistency across multiple viewpoints. These qualitative improvements clearly demonstrate the effectiveness and importance of our cross-view attention mechanism in achieving high-quality multi-view material estimation.

\subsubsection{Ablation on Feature Distillation Loss}

To evaluate the effectiveness of our feature distillation loss, we conduct an ablation study comparing model performance with and without feature distillation during the initial 50k training iterations. As shown in Table~\ref{tab:feat_distill}, incorporating the feature distillation loss consistently improves model performance across all evaluated metrics. Specifically, using feature distillation leads to significant gains in albedo estimation accuracy, increasing PSNR by 1.45 dB and SSIM by 0.007, while also improving metallic and roughness predictions by reducing their RMSE values. These results highlight the importance of feature distillation in effectively transferring visual knowledge between the albedo-optimized and material-specialized paths, ultimately enhancing overall material estimation quality.

\begin{table}[h!]
\centering
\begin{tabular}{lcccc}
\toprule
Method & Alb. PSNR & Alb. SSIM & Met. & Rough. \\ 
\midrule
W/ distillation loss & 26.64 & 0.897 & 0.070 & 0.073 \\
W/O distillation loss  & 25.19 & 0.890 & 0.072 & 0.079 \\
\bottomrule
\end{tabular}
\caption{Ablation study evaluating the effectiveness of the proposed feature distillation loss on PBR material estimation performance.}
\vskip -14px
\label{tab:feat_distill}
\end{table}

\subsubsection{Ablation on High-Resolution Processing}

\begin{figure}[H]
\centering
\includegraphics[width=\columnwidth]{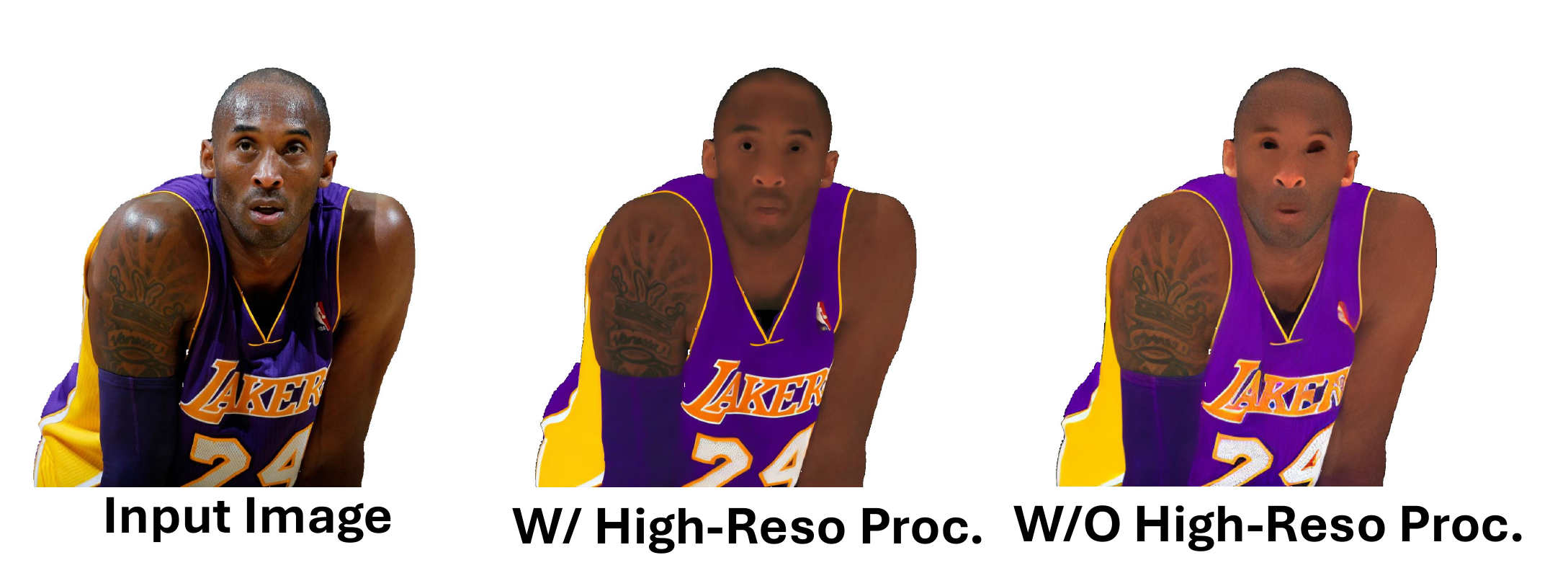}
\caption{Qualitative ablation study illustrating the impact of our proposed high-resolution processing strategy. }
\label{fig:aba_high_res}
\end{figure}

We present a qualitative ablation on our high-resolution processing strategy, as illustrated in \Fref{fig:aba_high_res}. When high-resolution processing is enabled, the model successfully preserves fine details and textures, closely matching the intricate features of the input image. In contrast, the version without high-resolution processing produces notably blurry and oversimplified results, losing essential facial and textural details. This ablation highlights the effectiveness and necessity of our high-resolution processing approach for maintaining detailed and realistic material predictions in practical high-resolution scenarios.

\begin{figure}[ht]
   \centering
   \includegraphics[width=0.5\textwidth]{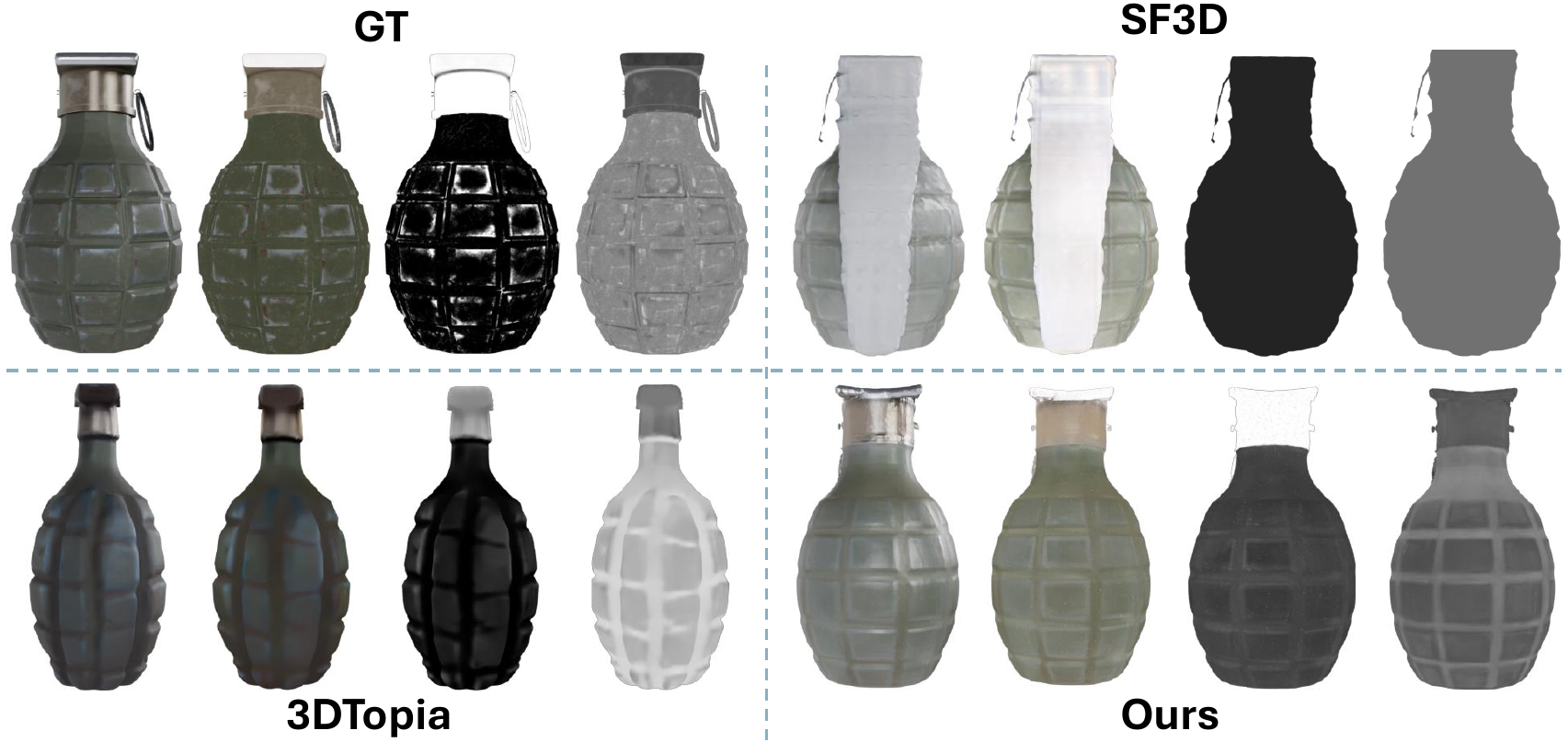} 
   \caption{Comparison with image-to-3D methods. Each row shows rendered RGB followed by estimated albedo, metallic, and roughness maps.}
   \label{fig:qual_result_3dgen}
\end{figure}

\subsection{Integration with Image-to-3D Pipelines}

\begin{figure}[ht]
   \centering
   \includegraphics[width=0.5\textwidth]{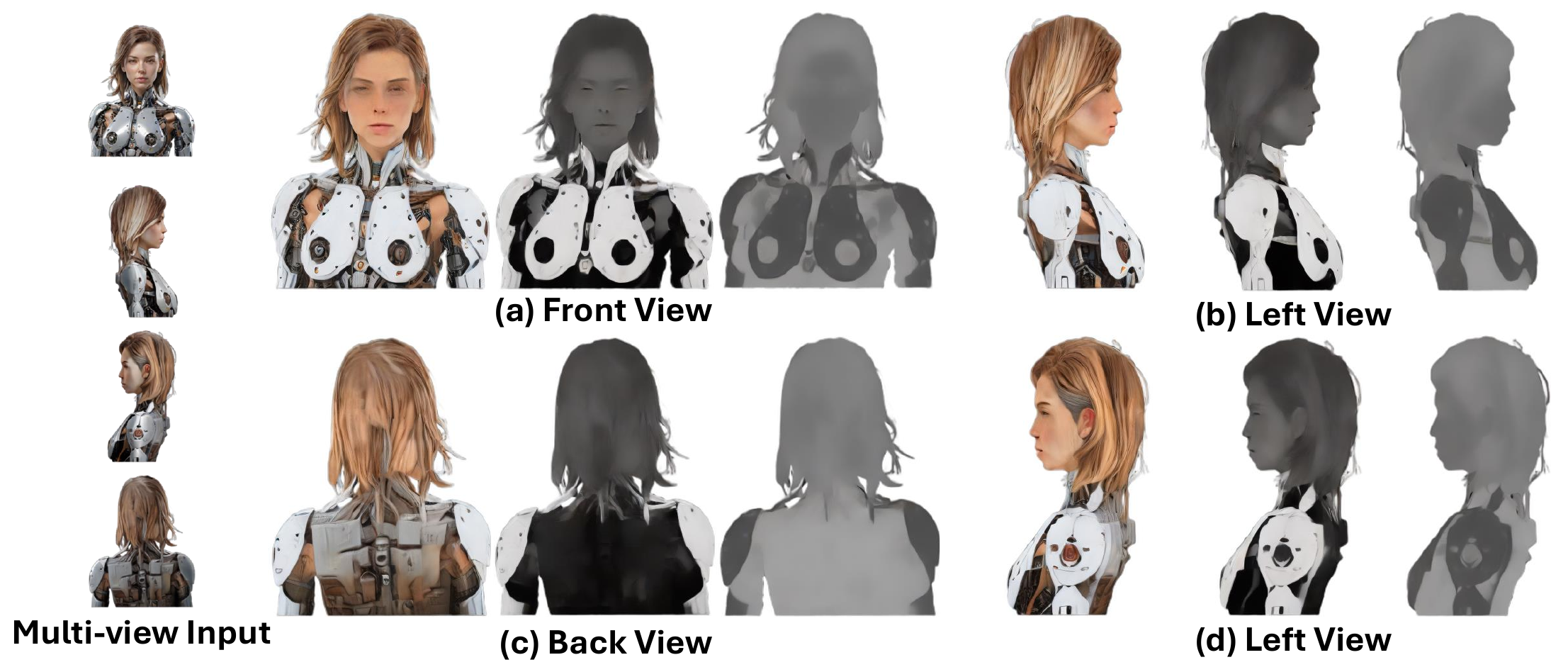}     
   \caption{Multi-view material estimation results. Given multi-view input images (left), our method produces consistent PBR decomposition across different viewpoints, showing \textbf{albedo}, \textbf{metallic}, and \textbf{roughness} maps for each view (a-d). Note how material properties remain coherent across viewpoint changes.}
   \label{fig:qual_result_multiview}
   \vspace{-0.3cm}
\end{figure}

\begin{figure}[ht]
  \centering
  \includegraphics[width=0.5\textwidth]{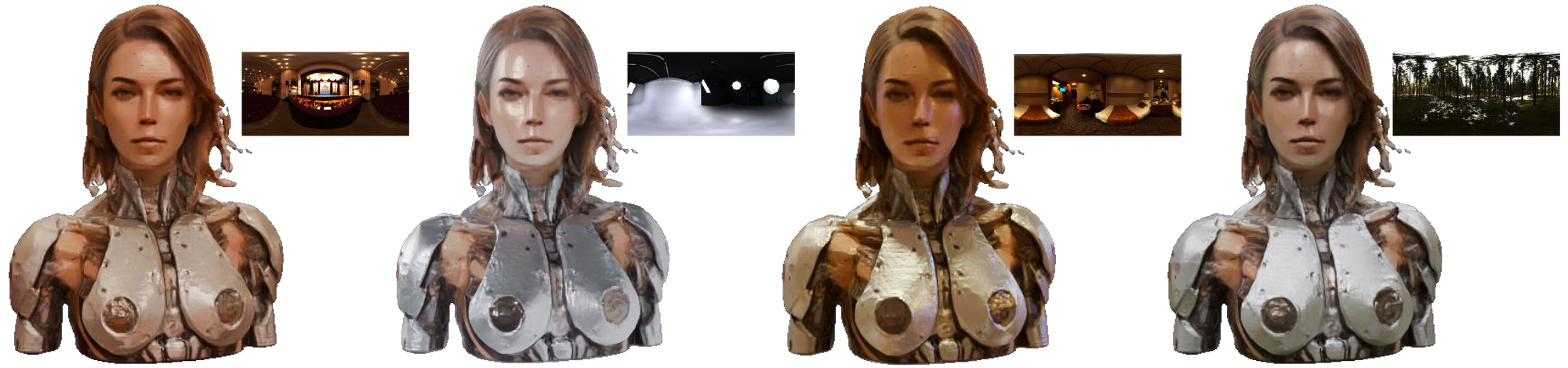}       
  \caption{Relighting results of our PBR-estimated textured mesh.}
  \label{fig:relighting}
  \vspace{-0.2cm}
\end{figure}

We demonstrate our method's effectiveness in enhancing modern image-to-3D generation pipelines by integrating it with Era3D~\cite{li2024era3d}. As shown in Figure~\ref{fig:qual_result_multiview}, our method first processes Era3D's multi-view outputs to obtain consistent PBR material properties across different viewpoints (front, left, right, back), which is crucial for high-quality 3D reconstruction. Using these consistent material estimates, we then generate a textured mesh following Era3D's pipeline with an additional PBR reconstruction loss. The quality of our estimated materials is validated through relighting experiments, as shown in Figure~\ref{fig:relighting}, where the reconstructed mesh maintains physically plausible appearance under various environment illuminations.

Furthermore, we compare our PBR estimation capabilities against other end-to-end image-to-3D methods that directly generate textured meshes with material properties, including SF3D~\cite{sf3d2024} and 3DTopia~\cite{chen2024primx}, as shown in \Fref{fig:qual_result_3dgen}. While these methods can generate textured 3D meshes, they often struggle to accurately decompose materials into physically-based properties that enable realistic relighting. In contrast, our method produces more precise PBR estimations that better support downstream rendering applications. Additional results and comparisons across diverse objects are provided in the supplementary material.

\section{Additional Results on In-the-Wild Images}
\begin{figure*}[ht]
\centering
\includegraphics[width=0.9\textwidth]{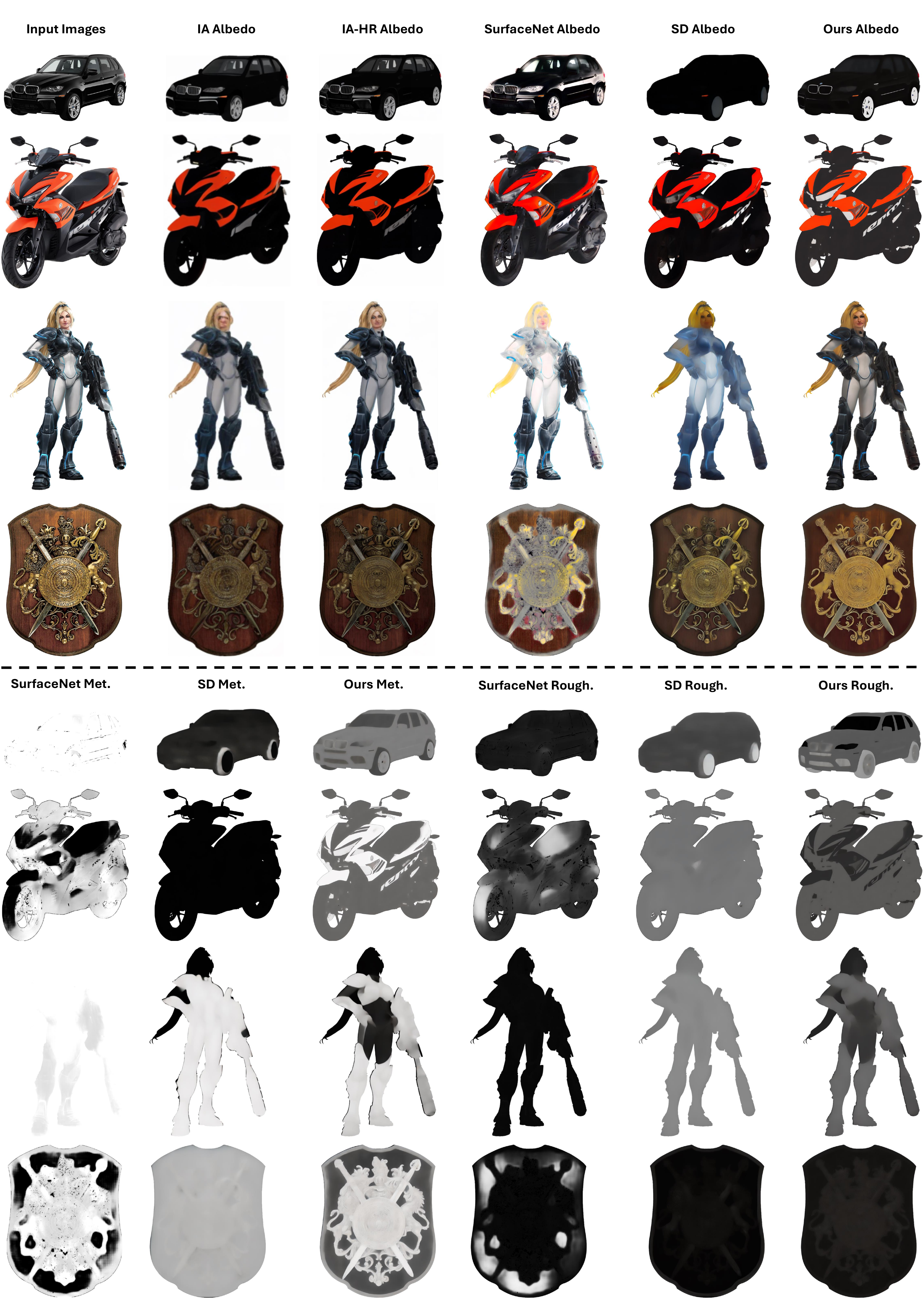}
\caption{Comparison on in-the-wild objects. }
\label{fig:sup_1}
\end{figure*}

\begin{figure*}[ht]
\centering
\includegraphics[width=1\textwidth]{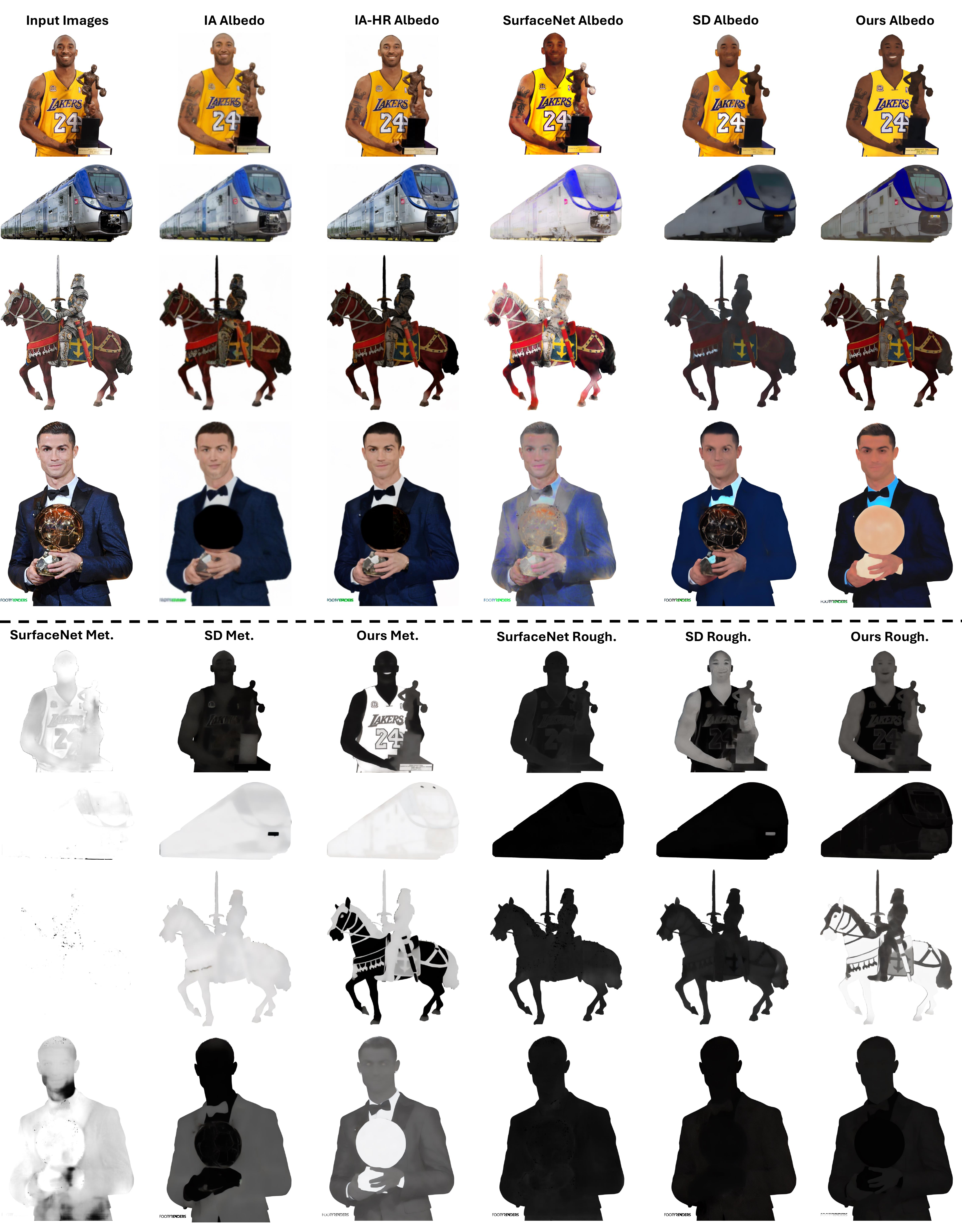}
\caption{Comparison on in-the-wild test objects. }
\label{fig:sup_2}
\end{figure*}

\begin{figure*}[ht]
\centering
\includegraphics[width=0.85\textwidth]{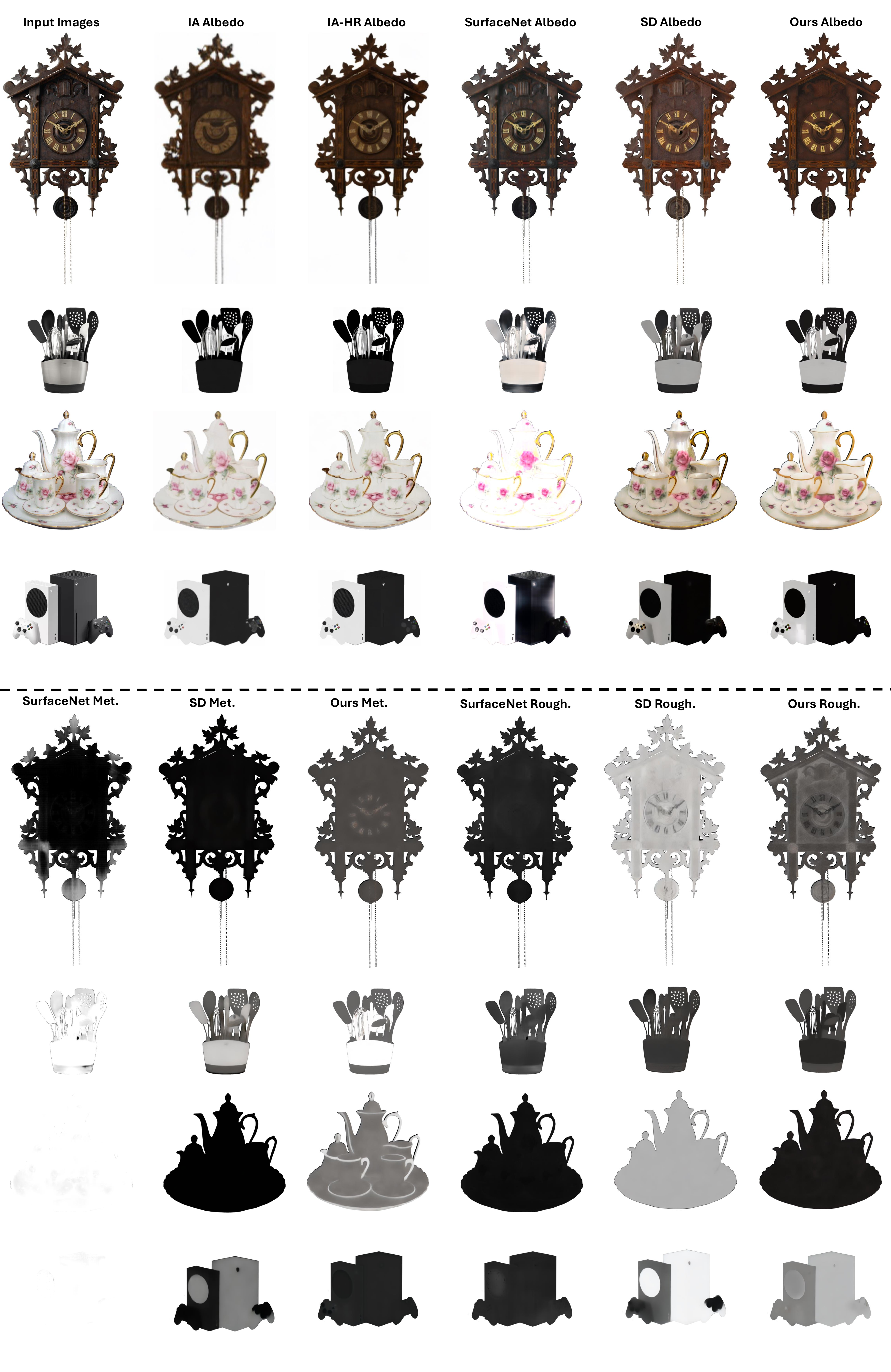}
\vspace{-0.5cm}
\caption{Comparison on in-the-wild test objects. }
\label{fig:sup_3}
\end{figure*}

Figures~\ref{fig:sup_1}--\ref{fig:sup_3} demonstrate our method's robust performance on diverse real-world objects with complex materials and lighting conditions, further validating our method's generalization capability beyond synthetic data.

\section{Conclusion}

We introduce DualMat, a novel dual-path diffusion framework for PBR material estimation that combines pretrained visual knowledge with specialized material understanding. Operating in complementary latent spaces for albedo and metallic-roughness estimation, our approach achieves state-of-the-art performance in material recovery from single images. Through feature distillation and efficient rectified flow sampling, we enable high-quality generation in just three steps, with extensions to high-resolution and multi-view scenarios. While the dual-path design increases memory cost, future work could explore more efficient architectures and language-guided material editing.

\bibliographystyle{ACM-Reference-Format}
\bibliography{sample-base}

\end{document}